\documentclass[10pt,twocolumn,letterpaper]{article}

\usepackage[pagenumbers]{cvpr}

\definecolor{cvprblue}{rgb}{0.21,0.49,0.74}
\usepackage[pagebackref,breaklinks,colorlinks,allcolors=cvprblue]{hyperref}
\usepackage{multirow}
\usepackage{utfsym}
\usepackage{makecell}

\usepackage[table]{xcolor}
\usepackage{tabularx}
\usepackage[dvipsnames]{xcolor}
\usepackage{marvosym}

\title{\makecell[c]{V-Loop: Visual Logical Loop Verification for Hallucination Detection\\in Medical Visual Question Answering}}

\author{Mengyuan Jin\footnotemark[1] \quad Zehui Liao\footnotemark[1] \quad Yong Xia\textsuperscript{\Letter} \\
Northwestern Polytechnical University\\
Xi'an 710072, China\\
{\tt\small \{myjin, merrical\}@mail.nwpu.edu.cn, yxia@nwpu.edu.cn}
}

\begin{document}
\maketitle

\renewcommand\thefootnote{\fnsymbol{footnote}}
\footnotetext{\textsuperscript{*}Equal Contribution \quad \textsuperscript{\Letter}Corresponding Author}

\begin{abstract}
Multimodal Large Language Models (MLLMs) have shown remarkable capability in assisting disease diagnosis in medical visual question answering (VQA).
However, their outputs remain vulnerable to hallucinations (i.e., responses that contradict visual facts), posing significant risks in high-stakes medical scenarios.
Recent introspective detection methods, particularly uncertainty-based approaches, offer computational efficiency but are fundamentally indirect, as they estimate predictive uncertainty for an image–question pair rather than verifying the factual correctness of a specific answer.
To address this limitation, we propose \textbf{Visual Logical Loop Verification (V-Loop)}, a training-free and plug-and-play framework for hallucination detection in medical VQA. 
V-Loop introduces a bidirectional reasoning process that forms a visually grounded logical loop to verify factual correctness. 
Given an input, the MLLM produces an answer for the primary input pair. 
V-Loop extracts semantic units from the primary QA pair, generates a verification question by conditioning on the answer unit to re-query the question unit, and enforces visual attention consistency to ensure answering both primary question and verification question rely on the same image evidence. 
If the verification answer matches the expected semantic content, the logical loop closes, indicating factual grounding; otherwise, the primary answer is flagged as hallucinated.
Extensive experiments on multiple medical VQA benchmarks and MLLMs show that V-Loop consistently outperforms existing introspective methods, remains highly efficient, and further boosts uncertainty-based approaches when used in combination.
\end{abstract}    
\section{Introduction}
\label{sec:intro}

\begin{figure}[t]
  \centering
   \includegraphics[width=0.95\linewidth]{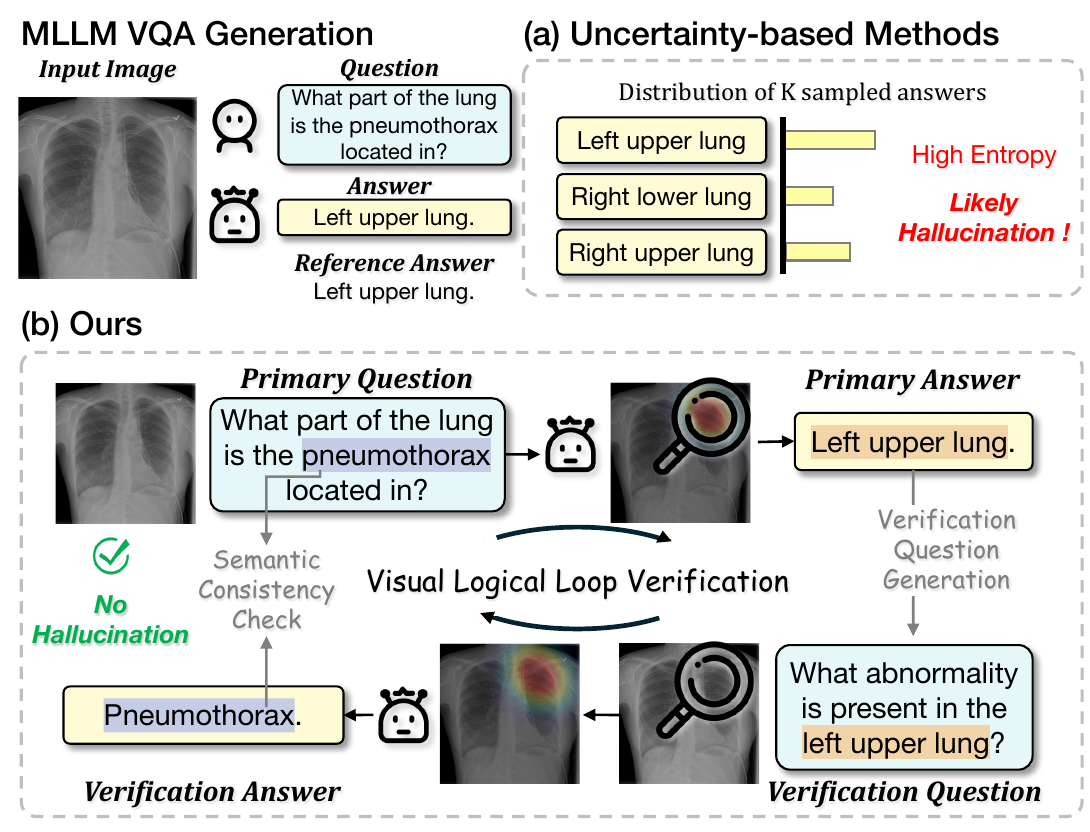}
   \caption{
   Comparison between uncertainty-based detection method and our proposed V-Loop.
   (a) Uncertainty-based methods detect hallucinations by sampling multiple answers and estimating semantic entropy.
   (b) V-Loop instead verifies the factual correctness of a specific answer through visual logical loop verification.
   }
   \label{fig:intro}
\end{figure}

Multimodal Large Language Models (MLLMs)~\cite{chen2024chexagent,li2023llava,chen2024towards,sellergren2025medgemma,xu2025lingshu} have shown strong capabilities in medical image understanding, integrating lesion recognition, severity assessment, and diagnostic reasoning within a unified framework. In medical visual question answering (VQA), they serve as promising clinical assistants that generate context-aware answers from image–question pairs. 
However, despite their impressive reasoning ability, medical MLLMs remain susceptible to \emph{hallucinations}, \textit{i.e.}, responses that contradict the visual facts grounded in the image~\cite{bai2024hallucination,liu2024survey}. In high-stakes clinical scenarios, hallucinated outputs can lead to misdiagnosis and loss of clinician trust, underscoring the urgent need for reliable hallucination detection~\cite{wu2024hallucination,zhu2025can,KhaBid_HallucinationAware_MICCAI2025}.

Existing hallucination detection methods can be broadly categorized by their external dependency.
\textbf{Collaborative approaches}~\cite{xiao2025detecting,Cohen2023LMVL,chen-etal-2024-complex} rely on auxiliary models or curated knowledge bases to verify generated answers, or they train dedicated detectors on carefully annotated hallucination data. 
While effective, their reliance on high-quality external resources limits generalization and scalability.
In contrast, \textbf{introspective approaches}~\cite{farquhar2024detecting,yu2024hallucidoctor,che2025eazy} operate solely within the MLLM, inferring hallucinations by estimating models' predictive uncertainty. As illustrated in Fig.~\ref{fig:intro}, these methods, such as Semantic Entropy (SE)~\cite{farquhar2024detecting} and VASE~\cite{Liao2025VASE}, assume that higher predictive entropy indicates a greater likelihood of hallucination. This paradigm is computationally efficient but remains indirect, as it assesses model uncertainty over an input image–question pair rather than verifying the factual correctness of a specific answer.

In real-world clinical practice, a lightweight, self-contained mechanism that internally verifies each generated answer is more deployable than resource-dependent or sampling-intensive methods.
Motivated by this gap, we propose \textbf{\textit{Visual Logical Loop Verification} (V-Loop)}, a training-free and plug-and-play framework for hallucination detection in medical VQA. 
V-Loop introduces a bidirectional reasoning process that constructs a visually grounded logical loop to assess factual consistency (see Fig.~\ref{fig:intro}). 
Given an image–question pair, the MLLM generates a \emph{primary answer} while recording its visual attention map. From this pair, two key semantic units (one from the question and one from the answer) are extracted to form the basis of verification. A \emph{verification question} is then constructed by conditioning on the answer’s semantic unit to re-query the question’s semantic unit, and the model is prompted to respond using the same visual evidence. If the verification answer semantically aligns with the expected content, the loop is closed, indicating that the original answer is visually grounded; otherwise, it is flagged as hallucinated. 
To further ensure that verification reasoning is grounded in the same visual evidence as primary reasoning, V-Loop enforces \textit{visual attention consistency} by requiring the MLLM to reuse its primary visual attention pattern when answering the verification question.

V-Loop offers three key advantages: (1) it is \textit{\textbf{training-free and model-agnostic}}, requiring no additional supervision or retraining; (2) it performs \textit{\textbf{visual-grounded self-verification}} in a single step, avoiding multi-sample uncertainty estimation; and (3) it achieves \textit{\textbf{strong generalization}} across diverse MLLMs and datasets. Extensive experiments on three medical VQA benchmarks and three MLLMs demonstrate that V-Loop consistently outperforms existing introspective and uncertainty-based methods. 

The main contributions of this work are threefold:
\begin{enumerate}
    \item We propose \textit{V-Loop}, a training-free, plug-and-play hallucination detection framework that verifies the factual correctness of medical VQA responses.

    \item  We design a \textit{visual logical loop verification mechanism} that verifies consistency between the model's bidirectional reasoning steps while maintaining \textit{visual attention consistency} between primary and verification stages.

    \item Comprehensive evaluations on three medical VQA benchmarks and three MLLMs show that V-Loop consistently outperforms competing methods and further improves uncertainty-based approaches when combined.
\end{enumerate}
\section{Related Work}
\label{sec:related}
\subsection{Medical VQA}

Medical VQA aims to generate clinically meaningful answers to image–question pairs~\cite{dong2025generative,LIN2023102611}. 
With the emergence of MLLMs~\cite{alayrac2022flamingo,liu2023llava,bai2025qwen2,zhu2025internvl3}, 
medical adaptations~\cite{li2023llava,xu2025lingshu,chen2024towards,sellergren2025medgemma,alkhaldi2024minigpt} 
have demonstrated remarkable progress in clinical image understanding through instruction tuning and domain-specific alignment. Despite these advances, hallucinations remain prevalent, as MLLMs often produce responses that deviate from visual evidence or clinical facts.

To evaluate the reliability of generated answers, traditional text similarity metrics such as BLEU~\cite{papineni-etal-2002-bleu}, ROUGE~\cite{lin-2004-rouge}, and BERTScore~\cite{Zhang*2020BERTScore:} primarily measure lexical overlap and fail to capture factual correctness or clinical semantics. To address this limitation, concept-aware metrics such as CHAIR~\cite{ren2020cgmvqa} and CBSS~\cite{talafha2018just} incorporate biomedical entity matching to evaluate responses at the concept or disease level. More recently, GREEN~\cite{ostmeier-etal-2024-green} introduced a fine-grained factuality metric that explicitly identifies clinical findings and semantic errors by comparing generated and reference answers. Following~\cite{Liao2025VASE}, we employ GREEN to automatically assign hallucination labels to MLLM outputs, enabling the construction of a standardized benchmark for hallucination detection in medical VQA.

\begin{figure*}[t]
  \centering
   \includegraphics[width=0.95\linewidth]{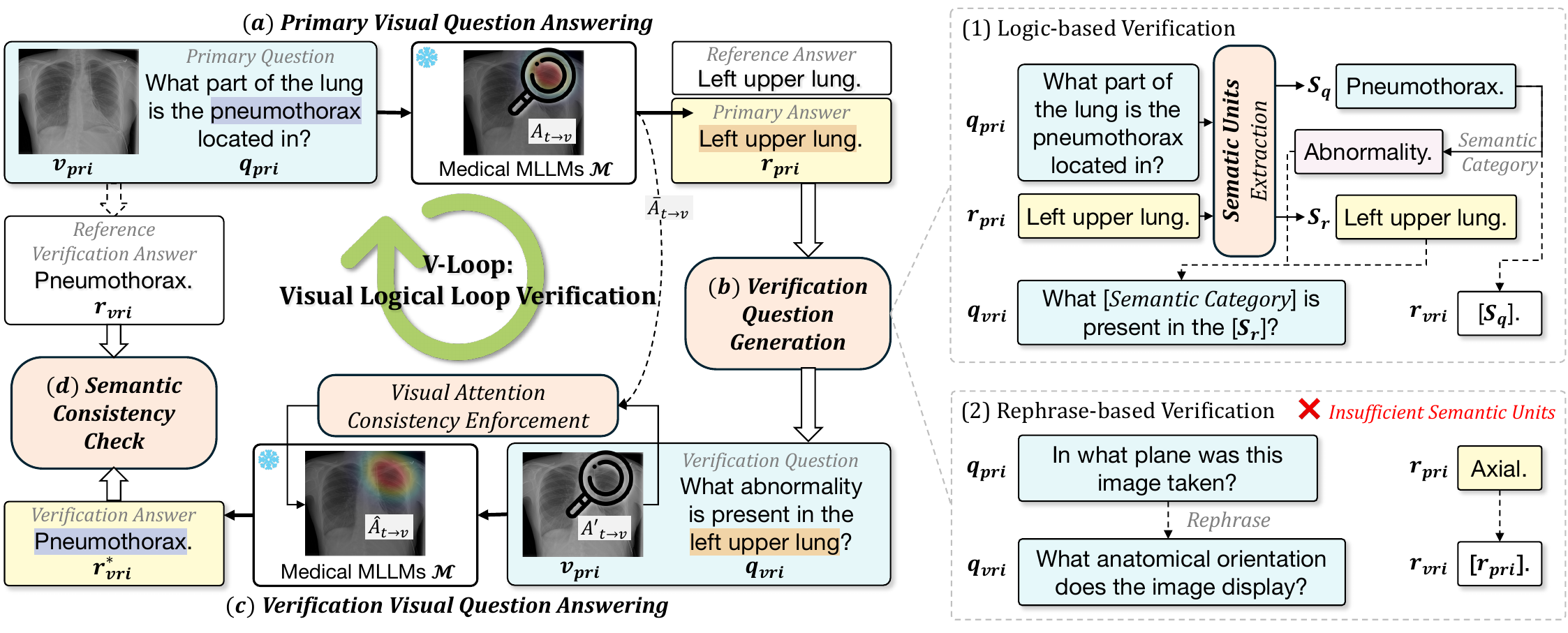}
   \caption{Overview of the proposed V-Loop framework.
    It consists of four stages, (a) primary visual question answering, where the MLLM generates the primary answer to the given image-question pair; (b) verification question generation, creating a verification question and its expected answer; (c) verification question answering, where the model answers the verification question under the same visual evidence; and (d) semantic consistency check, which evaluates the consistency between verification answer and its corresponding reference answer.}
   \label{fig:main}
\end{figure*}

\subsection{Hallucination Detection}

Hallucination detection methods can be broadly divided into two categories: \textbf{collaborative} and \textbf{introspective} frameworks.
\textbf{Collaborative approaches.}  
Collaborative methods rely on external models or knowledge sources to validate the correctness of generated responses. Some employ specialized \textit{detector networks}~\cite{xiao2025detecting,pmlr-v281-hardy25a,pandit2025teaching} trained on curated hallucination datasets to distinguish factual from erroneous outputs. Others adopt \textit{cross-model consistency} strategies~\cite{Cohen2023LMVL,zhang-etal-2023-sac3,yu2024hallucidoctor}, leveraging agreement among multiple LLMs or MLLMs to infer factual reliability. In addition, \textit{visual evidence verification} frameworks~\cite{sahu-etal-2024-pelican,10.1145/3689090.3689388} integrate expert vision models to align textual claims with visual content. Although effective, these approaches require high-quality detectors or external supervision, constraining their scalability and adaptability across domains.
\textbf{Introspective approaches.}  
Introspective methods eliminate external dependencies by enabling the MLLM to self-assess its factual consistency through internal reasoning or uncertainty estimation. Uncertainty-based methods~\cite{li-etal-2024-reference,farquhar2024detecting,Liao2025VASE} infer hallucinations by measuring predictive confidence or entropy, while \textit{self-consistency analysis}~\cite{wu-etal-2024-logical,zhang-etal-2023-sac3} evaluates semantic agreement across multiple reasoning paths. Despite their efficiency, these methods assess linguistic stability rather than factual grounding, and they do not verify whether a given answer aligns with visual evidence.

In contrast, our work introduces a visually grounded introspective framework that performs hallucination detection through a \textit{bidirectional reasoning loop}. By enforcing both semantic and visual attention consistency, the proposed method achieves reliable hallucination verification in a training-free, plug-and-play manner.
\section{Method}
Given a medical image-question pair $(q_{pri}, v_{pri})$, where $q_{pri}$ seeks specific information from $v_{pri}$, a medical MLLM $\mathcal{M}$ generates the response $r_{pri}$ in an auto-regressive manner. 
To distinguish it from the verification question-answer pair discussed later, we refer to $(q_{pri}, v_{pri})$ as the primary question-answer pair and $r_{pri}$ as the primary answer.
Our objective is to detect whether $r_{pri}$ contains hallucination.

\subsection{Overview}
As shown in Fig.~\ref{fig:main}, V-Loop proceeds in four compact steps:
(1) \textit{Primary VQA}: obtain $r_{pri}=\mathcal{M}(q_{pri},v_{pri})$ and record the model's text-to-image attention pattern during generation;
(2) \textit{Verification question generation (VQG)}: extract two semantic units from the primary triplet: $\mathcal{S}_q$ (from $q_{pri}$) and $\mathcal{S}_r$ (from $r_{pri}$), and produce a verification question $q_{vri}$ and its reference answer $r_{vri}$ using either logic-based inversion or rephrase-based equivalence;
(3) \textit{Verification VQA}: compute $r_{vri}^*=\mathcal{M}(q_{vri},v_{pri})$ while enforcing visual attention consistency to ensure that reasoning uses the same image evidence as the primary stage;
(4) \textit{Semantic consistency check}: compare $r_{vri}^*$ to $r_{vri}$ with a deterministic semantic evaluator $\mathcal{E}$: disagreement indicates a broken loop and signals a hallucination.
We now delve into the details.

\subsection{Primary VQA}
During Primary VQA, the medical MLLM $\mathcal{M}$ generates a response $r_{pri}$ based on the given image–question pair $(q_{pri}, v_{pri})$. Then the concatenation of $q_{pri}$ and $r_{pri}$ forms a primary claim.
For example, given the question `What part of the lung is the pneumothorax located in?' and the answer `Left upper lung,' the resulting claim becomes `The pneumothorax is located in the left upper lung.'
From the primary claim, we extract two compact \textbf{\textit{semantic units}}: $\mathcal{S}_q$ is extracted from $q_{pri}$, and $\mathcal{S}_r$ is extracted from $r_{pri}$. Each semantic unit may represent a lesion, anatomical region, attribute, or other clinical concept, depending on the content of the primary pair. In the example above, $\mathcal{S}_{q}$ is `pneumothorax' and $\mathcal{S}_{r}$ is `left upper lung'.

\subsection{Verification Question Generation}
\label{subsection:vqg}
Intuitively, the \textit{verification question} is conditioned on the semantic unit $\mathcal{S}_{r}$ extracted from the primary answer and is designed to re-query the semantic unit $\mathcal{S}_{q}$ originating from the primary question.  
The expected verification answer is therefore the semantic unit $\mathcal{S}_{q}$ itself, forming a semantic alignment between the primary and verification stages.
Some question-answer pairs naturally contain two distinct semantic units, such as a queried entity and its contextual attribute, which enables the construction of a logical loop. However, others involve only a single semantic focus or lack a clear role exchange (\eg, `What imaging modality is used?'$\rightarrow$`CT'), making such loop-based verification infeasible.
To handle both cases effectively, we design two complementary strategies for verification question generation: 
(1) \textit{logic-based verification}, which performs bidirectional reasoning consistency verification when dual semantic units are available, and 
(2) \textit{rephrase-based verification}, which focuses on semantic equivalence checking when a logical loop cannot be formed.

\textbf{Logic-based Verification.}  
When both primary question and primary answer contain distinct semantic units $\mathcal{S}_q$ and $\mathcal{S}_r$, we use $\mathcal{S}_r$ as the inquiry condition to re-query $\mathcal{S}_q$ during the verification stage.
To enhance clarity and clinical relevance, $\mathcal{S}_q$ is further annotated with its \textit{semantic category} (\eg, `Abnormality', `Organ', or `Attribute'), which provides the language model with explicit contextual guidance.  
This category-level supervision encourages the model to generate verification questions that are semantically precise, clinically grounded, and aligned with the original diagnostic intent.  
For example, when $\mathcal{S}_q$ (\eg, `pneumothorax') is labeled as `Abnormality', the model is prompted to generate domain-consistent questions such as `What abnormality is located in [region]?' rather than a generic or ambiguous one like `What is located in [region]?'.

\textbf{Rephrase-based verification.}  
Not all primary question-answer pairs support the construction of a logic-based verification question. Suppose the pair lacks clear dual semantic units (single-focus queries such as modality, plane, or general image properties). In that case, we reformulate the primary question $q_{pri}$ into a semantically equivalent question $q_{vri}$ that preserves its intent while varying its linguistic structure, and then let the reference answer be the original one, \ie, $r_{vri}=r_{pri}$. This reduces V-Loop to a semantic-equivalence check when logical inversion cannot be formed.

\begin{figure}[t]
  \centering
   \includegraphics[width=0.98\linewidth]{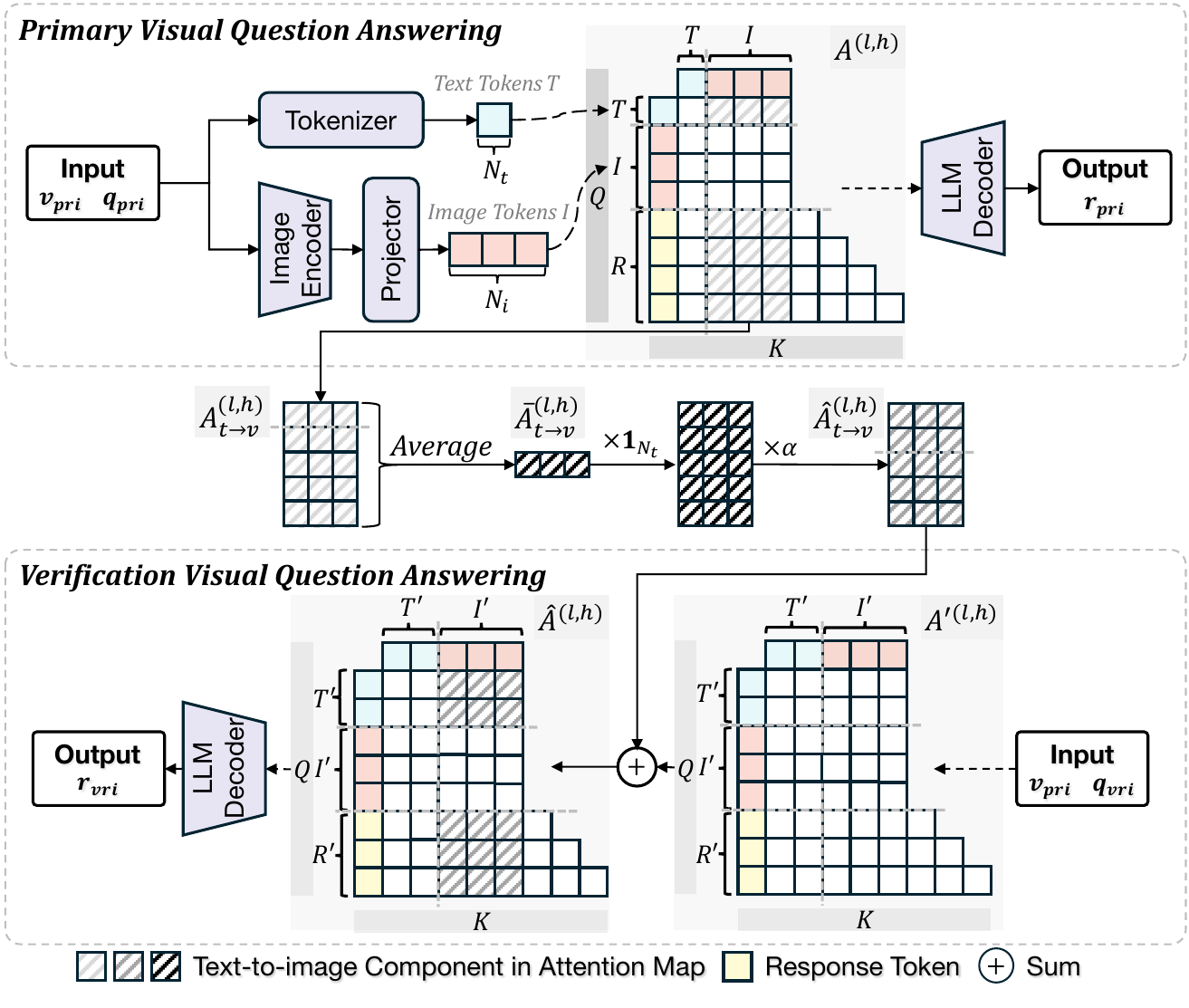}
   \caption{Visual Attention Consistency.
   The upper panel shows the primary stage, and the lower shows the verification process.}
   \label{fig:echogen}
\end{figure}

\subsection{Verification VQA}
\label{subsec:backward}
In the verification stage, the generated verification question $q_{vri}$ and the original image $v_{pri}$ are fed into the MLLM $\mathcal{M}$ to produce the verification answer $r_{vri}^*$. While this process mirrors the primary VQA stage, we introduce a \textit{\textbf{Visual Attention Consistency (VAC)}} mechanism. This constraint ensures that the model performs verification based on the \textbf{\textit{same visual evidence}} it relied upon when answering the primary question. Without this, the model might attend to unrelated image regions during the verification stage, leading to spurious results that are visually inconsistent with its original prediction. 
The VAC mechanism forces the MLLM to reuse the visual attention patterns from the primary answering stage, thereby maintaining consistent image-level evidence across both processes.

\textbf{Attention aggregation.}  
In the primary reasoning stage, let the attention map from the $h$-th head in the $l$-th layer be $A^{(l,h)}$. We isolate the text-to-image component, $A_{t \rightarrow v}^{(l,h)} \in \mathbb{R}^{N_t \times N_v}$, which captures the attention from text tokens (originating from the primary question and its generated answer) to the $N_v$ visual tokens. $N_t$ is the number of text tokens. To create a global summary of visual focus, we average these text-to-image attention weights across all $N_t$ text tokens, all $H$ heads, and all $L$ layers. This yields a single aggregated visual attention vector:
\begin{equation}
  \bar{A}_{t \rightarrow v} = \frac{1}{L H N_t} \sum_{l=1}^{L}\sum_{h=1}^{H}\sum_{t=1}^{N_t} A_{t \rightarrow v}^{(l,h)} \in \mathbb{R}^{N_v},
  \label{eq:attn_weight_text_to_image_average}
\end{equation}
which represents the model's overall attention pattern on image tokens from the primary reasoning stage.

\textbf{Applying Aggregated Attention to Verification.} 
In the backward verification stage, this aggregated visual pattern $\bar{A}_{t \rightarrow v}$ is utilized to regularize the verification attention. We first denote the attention map of the verification stage as ${A^{\prime}}^{(l,h)}$, and scale $\bar{A}_{t \rightarrow v}$ by a coefficient $\alpha$ before injecting it into the text-to-image attention component ${A^{\prime}}_{t \rightarrow v}^{(l,h)}$:
\begin{equation}
    \hat{A}_{t \rightarrow v}^{(l,h)} = {A^{\prime}}_{t \rightarrow v}^{(l,h)} + \alpha \times (\mathbf{1}_{N_t} \cdot \bar{A}_{t \rightarrow v}) \in \mathbb{R}^{N_t \times N_v},
  \label{eq:attn_weight_reweight}
\end{equation}
where $\mathbf{1}_{N_t} \in \mathbb{R}^{N_t \times 1}$ is an all-ones vector used to broadcast $\bar{A}_{t \rightarrow v}$ across all $N_t$ text tokens. 

The resulting reweighted block $\hat{A}_{t \rightarrow v}^{(l,h)}$ is substituted back into the full attention matrix ${A^{\prime}}^{(l,h)}$, yielding the modified full matrix $\hat{A}^{(l,h)}$.
To restore the probabilistic nature of the attention weights (which was disrupted by the addition), we apply a row-wise softmax operation to each row of $\hat{A}^{(l,h)}$.
The normalized attention map $\tilde{A}^{(l,h)}$ is then used in the standard attention computation (\ie, multiplied by the value matrix $V$) to produce the layer's output. This VAC constraint guides the MLLM to perform verification using the same image evidence it relied upon during primary reasoning, thereby enforcing a visually grounded logical loop.

\subsection{Semantic Consistency Check}
We assess the semantic consistency between the generated verification answer $r_{vri}^*$ and its reference answer $r_{vri}$. 
Note that the reference answer $r_{vri}$ is defined as $\mathcal{S}_q$ for logic-based verification or $r_{pri}$ for rephrase-based verification. 
This assessment determines whether the model forms a \textbf{logically closed reasoning loop}. 
If the two answers are semantically aligned, the primary claim is deemed grounded, indicating minimal hallucination risk.
Conversely, a mismatch suggests the primary answer is hallucinated.

To assess this alignment, we utilize an auxiliary LLM, DeepSeek-V3.2-Exp~\cite{deepseekai2024deepseekv32}, as a deterministic semantic evaluator, denoted as $\mathcal{E}$, to provide a reliable judgment.
Given a pair $(r_{vri}^*,r_{vri})$, the evaluator returns a similarity score $s \in [0,1]$ measuring their semantic agreement: 
\begin{equation}
s = \mathcal{E}(r_{vri}^*, r_{vri}).
\end{equation}
A prediction is considered consistent only if $s = 1.0$. To ensure deterministic and reproducible outputs, this evaluation is conducted using a fixed prompt template with the evaluator's generation temperature set to $0.0$.
\section{Experiments}
\subsection{Experimental Setup}

\begin{table*}[t]
  \centering
  \caption{
  Performance (AUC (\%) and AUG (\%)) of V-Loop and seven competing methods. The best and second-best results are highlighted in \textbf{bold} and \underline{underline}, respectively. * represents method which is uncertainty-based and requires additional $K$ times response generation.
  }
  \setlength\tabcolsep{6pt}
  \begin{tabular}{l|cc|cc|cc|cc|cc|cc}
    \toprule
    \multirow{3}{*}{Method} & \multicolumn{4}{c|}{VQA-RAD} & \multicolumn{4}{c|}{VQA-Med-2019} & \multicolumn{4}{c}{SLAKE} \\
    \cline{2-13}
     & \multicolumn{2}{c|}{Open-Ended} & \multicolumn{2}{c|}{All} & \multicolumn{2}{c|}{Open-Ended} & \multicolumn{2}{c|}{All} & \multicolumn{2}{c|}{Open-Ended} & \multicolumn{2}{c}{All} \\
    \cline{2-13}
     & AUC & AUG & AUC & AUG & AUC & AUG & AUC & AUG & AUC & AUG & AUC & AUG \\
    \bottomrule
    \multicolumn{13}{c}{MedGemma-4B-it~\cite{sellergren2025medgemma}} \\
    \toprule
    AvgProb~\cite{li-etal-2024-reference}
    & 36.36 & 43.47 & 44.49 & 54.25 
    & 42.96 & 35.43 & 44.41 & 38.33
    & 44.02 & 58.77 & 43.01 & 56.26 \\
    AvgEnt~\cite{li-etal-2024-reference}
    & \underline{63.50} & 60.17 & 55.60 & 58.33
    & 56.94 & 32.85 & 55.42 & 35.40 
    & 55.98 & 60.72 & 56.98 & 57.20 \\
    MaxProb~\cite{li-etal-2024-reference}
    & 36.86 & 45.37 & 43.71 & 53.12
    & 41.28 & 30.00 & 43.64 & 35.69
    & 43.26 & 56.83 & 42.65 & 55.02 \\
    MaxEnt~\cite{li-etal-2024-reference}
    & 62.80 & 60.17 & 56.11 & 58.97 
    & 58.58 & 33.58 & 56.42 & 34.84
    & 56.59 & 60.89 & 57.45 & 57.23 \\
    SE*~\cite{farquhar2024detecting}
    & 63.33 & \underline{63.03} & 56.65 & \underline{59.76}
    & 61.88 & \underline{47.02} & 60.91 & 48.78
    & \underline{60.90} & \underline{68.42} & 56.84 & \underline{59.62} \\
    RadFlag*~\cite{pmlr-v259-zhang25c}
    & 62.91 & 62.87 & \underline{58.69} & 57.94 
    & \underline{64.91} & 36.42 & \underline{62.95} & \underline{50.14}
    & 59.18 & 63.57 & \textbf{59.56} & 58.53 \\
    \rowcolor{gray!10}
    Ours
    & \textbf{66.32} & \textbf{66.51} & \textbf{60.30} & \textbf{66.91} 
    & \textbf{66.02} & \textbf{50.92} & \textbf{63.28} & \textbf{52.15} 
    & \textbf{61.87} & \textbf{68.60} & \underline{59.31} & \textbf{65.15} \\
    \bottomrule
    \multicolumn{13}{c}{Lingshu-7B~\cite{xu2025lingshu}} \\
    \toprule
    AvgProb~\cite{li-etal-2024-reference}
    & 50.52 & 47.85 & 48.27 & 56.36 
    & 45.93 & \underline{46.15} & 45.64 & \underline{48.61}
    & 46.87 & 70.07 & 46.75 & 69.97 \\
    AvgEnt~\cite{li-etal-2024-reference}
    & 49.48 & 47.07 & \underline{51.73} & \underline{60.96}
    & 54.07 & 39.98 & 54.36 & 41.31
    & 53.13 & 74.38 & \underline{53.25} & \underline{75.65} \\
    MaxProb~\cite{li-etal-2024-reference}
    & 50.37 & 47.69 & 48.24 & 55.44 
    & 45.87 & 44.81 & 45.60 & 47.82
    & 46.84 & 69.82 & 46.75 & 70.09 \\
    MaxEnt~\cite{li-etal-2024-reference} 
    & 49.63 & 47.13 & \textbf{51.76} & \underline{60.96} 
    & 54.13 & 40.00 & 54.40 & 41.30 
    & 53.16 & 74.39 & \underline{53.25} & \underline{75.65} \\
    SE*~\cite{farquhar2024detecting}
    & \underline{52.40} & \underline{49.64} & 50.96 & 59.16
    & 51.80 & 39.39 & 51.90 & 38.72 
    & 50.58 & 73.61 & 50.89 & 75.61 \\
    RadFlag*~\cite{pmlr-v259-zhang25c}
    & 51.90 & 48.35 & 51.55 & 59.29
    & \underline{55.24} & 40.67 & \underline{55.91} & 42.77 
    & \underline{53.29} & \underline{75.05} & 52.52 & 75.44 \\
    \rowcolor{gray!10}
    Ours
    & \textbf{70.00} & \textbf{56.47} & 50.04 & \textbf{62.84} 
    & \textbf{71.93} & \textbf{66.33} & \textbf{65.66} & \textbf{65.03}
    & \textbf{67.68} & \textbf{84.01} & \textbf{59.79} & \textbf{81.83} \\
    \bottomrule
    \multicolumn{13}{c}{InternVL3-8B~\cite{li2023llava}} \\
    \toprule
    AvgProb~\cite{li-etal-2024-reference}
    & 39.88 & 42.12 & 39.77 & 43.33 
    & 38.89 & 33.64 & 39.74 & 37.45 
    & 41.62 & 46.91 & 41.24 & 46.19 \\
    AvgEnt~\cite{li-etal-2024-reference}
    & 58.89 & 51.44 & 59.67 & 54.34 
    & 60.46 & 42.05 & 60.17 & 47.90 
    & 58.47 & 54.34 & 58.09 & 55.58 \\
    MaxProb~\cite{li-etal-2024-reference}
    & 38.40 & 38.05 & 38.66 & 40.72 
    & 36.44 & 28.27 & 38.09 & 32.09 
    & 40.68 & 45.28 & 40.38 & 44.39 \\
    MaxEnt~\cite{li-etal-2024-reference} 
    & 61.17 & 52.11 & \underline{61.59} & 57.14 
    & 63.81 & 46.94 & 62.58 & 54.87 
    & 59.53 & 54.51 & 58.81 & 55.93 \\
    SE*~\cite{farquhar2024detecting}
    & 61.26 & 53.89 & 60.04 & 56.01 
    & 66.18 & 52.19 & 66.35 & 46.71 
    & 58.24 & 54.03 & \underline{60.61} & \underline{58.31} \\
    RadFlag*~\cite{pmlr-v259-zhang25c} 
    & \underline{67.06} & \underline{55.91} & \textbf{63.99} & 57.98 
    & \underline{67.69} & \underline{53.58} & \underline{70.25} & \underline{55.28} 
    & \textbf{64.93} & 58.76 & \textbf{62.12} & \textbf{59.75} \\
    \rowcolor{gray!10}
    Ours
    & \textbf{68.20} & \textbf{57.68} & 60.98 & \textbf{58.15} 
    & \textbf{72.86} & \textbf{56.58} & \textbf{70.55} & \textbf{55.82} 
    & \underline{59.97} & \textbf{59.77} & 56.04 & 57.96 \\
    \bottomrule
  \end{tabular}
  \label{tab:main_result}
\end{table*}

\textbf{Datasets.}
We evaluated \textit{V-Loop} against competing hallucination detection methods on three medical VQA benchmarks that contain both closed-ended (binary \textit{`yes'}/\textit{`no'}) and open-ended question-answer pairs:
\textbf{VQA-RAD}~\cite{lau2018dataset} is a manually curated and clinically verified radiology VQA dataset comprising 2,244 QA pairs over 314 medical images. Its test set consists of 451 pairs from 203 images, including 200 open-ended and 251 closed-ended questions.
\textbf{VQA-Med-2019}~\cite{abacha2019vqa}, introduced in the ImageCLEF 2019 VQA-Med challenge, contains 4,200 radiology images with 15,292 QA pairs. The test set includes 500 questions on 500 images across four categories,  modality, plane, organ system, and abnormality, of which 436 are open-ended and 64 are closed-ended.
\textbf{SLAKE}~\cite{liu2021slake} is a semantically annotated bilingual dataset with 14,028 QA pairs over 642 images. We use the English portion of the test set (2,094 total, 645 open-ended / 416 closed-ended).
All reported results include both the open-ended subset (most informative for hallucination diagnosis) and the full test set.

\textbf{Backbones.} 
We evaluated across two medical MLLMs and a general-purpose MLLM, which cover the common tradeoffs between medical specialization and model capacity.
\textbf{MedGemma-4B-it}~\cite{sellergren2025medgemma} is built upon the Gemma3~\cite{team2025gemma} architecture with a SigLIP-based visual encoder and is further aligned through large-scale medical image–text pre-training. 
\textbf{Lingshu-7B}~\cite{xu2025lingshu} is a Qwen2.5-VL-based~\cite{bai2025qwen2} multimodal model tailored for medical vision–language understanding through staged alignment and instruction tuning. 
\textbf{InternVL3-8B}~\cite{zhu2025internvl3} follows the `ViT–MLP–LLM' paradigm, integrating a Qwen2.5 series and InternLM3-8B language model with InternViT vision encoders and an MLP projector to bridge modalities.

\textbf{Hallucination Detection Benchmark Construction.}
Originally, the medical VQA datasets cannot be directly used to evaluate hallucination detection, as they provide neither the model-generated responses to be assessed nor the necessary hallucination labels.
Following prior work~\cite{Liao2025VASE}, we (i) generated answers for every test sample using each evaluated MLLM and (ii) computed hallucination labels automatically via the \textbf{GREEN} model~\cite{ostmeier-etal-2024-green}. 
GREEN compares a generated answer to the reference, identifies matched findings and clinical errors, and returns
\begin{equation}
GREEN = \frac{\# \  matched\ findings}{\# \ matched\ findings\ +\ \# \ errors}.
\end{equation}
We treat samples with $\text{GREEN}<1.0$ as containing hallucinations ($\text{label} = 1.0$) and $\text{GREEN}=1.0$ as non-hallucinated ($\text{label} = 0.0$). 
This automatic evaluation pipeline facilitates scalable benchmarking of hallucination detection methods in line with recent practices.

\textbf{Evaluation Metrics.} 
Following prior hallucination detection studies~\cite{farquhar2024detecting,Liao2025VASE}, we reported (1) the area under the ROC curve (AUC) and (2) the area under the GREEN Curve (AUG). 
\textbf{AUC} reflects the likelihood that a randomly chosen hallucinated response receives a higher detection score than a randomly chosen non-hallucinated one. For \textbf{AUG}, we first defined the \textit{mean GREEN score at X\%} as the average GREEN score of the top X\% most confident samples, as ranked by each uncertainty-based method being evaluated. The AUG value is subsequently obtained by calculating the total area under the resulting confidence curve. This curve is formed by plotting the Mean GREEN score (y-axis) against $X\%$ (x-axis), by integrating over $X$ ranging from 1\% to 100\% (with a 1\% step size). This process effectively sums the mean scores across the entire confidence range.
Larger AUC and AUG values consistently indicate a stronger hallucination detection capability.

\textbf{Baselines.}  
We compared our \textit{V-Loop} framework against seven baselines, categorized by their approach to uncertainty estimation.
Four baselines rely on MLLM's output token probabilities. 
AvgProb~\cite{li-etal-2024-reference} and MaxProb~\cite{li-etal-2024-reference} measure model confidence by computing the average or maximum token probability of the generated sequence.
AvgEnt~\cite{li-etal-2024-reference} and MaxEnt~\cite{li-etal-2024-reference} estimate model uncertainty by calculating the average or maximum entropy over the token-level probability distribution.
Other baselines assess uncertainty by generating multiple responses.
SE~\cite{farquhar2024detecting} estimates predictive uncertainty at the semantic level by generating multiple responses and computing the entropy of the semantic predictive distribution.
VASE~\cite{Liao2025VASE} further enhances SE by amplifying the influence of visual evidence during semantic entropy estimation. 
RadFlag~\cite{pmlr-v259-zhang25c} is a consistency-based approach that quantifies the degree of agreement among multiple generated answers.

For the uncertainty-based methods (SE and RadFlag), the response generation time is set to 2.
Please note that VASE estimates semantic entropy for both the original and perturbed images, resulting in a sampling frequency that is twice that of SE and RadFlag.
Given its prohibitively high computational cost and to ensure a fair comparison, VASE is excluded from the main comparative experiments.

\begin{table*}[t]
  \centering
  \caption{Performance (AUC (\%) and AUG (\%)) of uncertainty-based methods and their V-Loop-enhanced variants on VQA-Med-2019.}
  \setlength\tabcolsep{3pt}
  \begin{tabular}{l|cc|cc|cc|cc|cc|cc}
    \toprule
     \multirow{3}{*}{Method} & \multicolumn{2}{c|}{Open-Ended} & \multicolumn{2}{c|}{All} & \multicolumn{2}{c|}{Open-Ended} & \multicolumn{2}{c|}{All} & \multicolumn{2}{c|}{Open-Ended} & \multicolumn{2}{c}{All} \\
    \cline{2-13}
     & AUC & AUG & AUC & AUG & AUC & AUG & AUC & AUG & AUC & AUG & AUC & AUG \\
    \cline{2-13}
    & \multicolumn{4}{c|}{MedGemma-4B-it~\cite{sellergren2025medgemma}} & \multicolumn{4}{c|}{Lingshu-7B~\cite{xu2025lingshu}} & \multicolumn{4}{c}{InternVL3-8B~\cite{li2023llava}} \\
    \bottomrule
    SE~\cite{farquhar2024detecting} 
    & 61.88 & 47.02 & 60.91 & 48.78
    & 51.80 & 39.39 & 51.90 & 38.72
    & 66.18 & 52.19 & 66.35 & 46.71 \\
    \rowcolor{gray!10}
    \ \ \ \ \ $+$ Ours 
    & 69.02 & 51.93 & 66.07 & 53.77 
    & 72.51 & 66.33 & 66.01 & 65.62
    & 75.60 & 58.99 & 74.14 & 57.67 \\ 
    \ \ \ \ \ \ \ \ \ $\Delta$ 
    & \color{ForestGreen}$+$7.14 & \color{ForestGreen}$+$4.91 & \color{ForestGreen}$+$5.16 & \color{ForestGreen}$+$4.99 
    & \color{ForestGreen}$+$20.71 & \color{ForestGreen}$+$26.94 & \color{ForestGreen}$+$14.11 & \color{ForestGreen}$+$26.90
    & \color{ForestGreen}$+$9.42 & \color{ForestGreen}$+$6.80 & \color{ForestGreen}$+$7.79 & \color{ForestGreen}$+$10.96 \\ \hline
    RadFlag~\cite{pmlr-v259-zhang25c}  
    & 64.91 & 36.42 & 62.95 & 50.14
    & 55.24 & 40.67 & 55.91 & 42.77 
    & 67.69 & 53.58 & 70.25 & 55.28 \\
    \rowcolor{gray!10}
    \ \ \ \ \ $+$ Ours 
    & 71.11 & 53.89 & 67.02 & 53.91 
    & 73.42 & 66.43 & 67.23 & 65.22
    & 76.14 & 58.23 & 75.92 & 60.79 \\
    \ \ \ \ \ \ \ \ \ $\Delta$ 
    & \color{ForestGreen}$+$6.20 & \color{ForestGreen}$+$17.47 & \color{ForestGreen}$+$4.07 & \color{ForestGreen}$+$3.77 
    & \color{ForestGreen}$+$18.18 & \color{ForestGreen}$+$25.76 & \color{ForestGreen}$+$11.32 & \color{ForestGreen}$+$22.45
    & \color{ForestGreen}$+$8.45 & \color{ForestGreen}$+$4.65 & \color{ForestGreen}$+$5.67 & \color{ForestGreen}$+$5.51 \\ \hline
    VASE~\cite{Liao2025VASE} 
    & 70.92 & 51.46 & 68.19 & 44.71 
    & 68.02 & 61.23 & 67.09 & 57.33
    & 72.35 & 54.67 & 72.07 & 54.18 \\
    \rowcolor{gray!10}
    \ \ \ \ \ $+$ Ours 
    & \textbf{73.51} & \textbf{54.67} & \textbf{69.56} & \textbf{54.90}
    & \textbf{76.01} & \textbf{68.60} & \textbf{68.80} & \textbf{68.43}
    & \textbf{78.14} & \textbf{61.57} & \textbf{76.10} & \textbf{61.76} \\
    \ \ \ \ \ \ \ \ \ $\Delta$ 
    & \color{ForestGreen}$+$2.59 & \color{ForestGreen}$+$3.21 & \color{ForestGreen}$+$1.37 & \color{ForestGreen}$+$10.19 
    & \color{ForestGreen}$+$7.99 & \color{ForestGreen}$+$7.37 & \color{ForestGreen}$+$1.71 & \color{ForestGreen}$+$11.10
    & \color{ForestGreen}$+$5.79 & \color{ForestGreen}$+$6.90 & \color{ForestGreen}$+$4.03 & \color{ForestGreen}$+$7.58 \\
    \bottomrule
  \end{tabular}
  \label{tab:complement}
\end{table*}

\textbf{Implementation Details.}  
Primary answers from all MLLMs were generated with decoding temperature $\textit{T}=0.1$. For sampling-based baselines (SE, RadFlag, VASE) we used $\textit{T}=1.0$ during the sampling stage. In V-Loop the verification answer is generated with $\textit{T}=0.1$. The attention reweighting coefficient is set to $\alpha = 0.7$ unless otherwise noted. All experiments were run on a single NVIDIA GeForce RTX 4090D GPU, and reported results are aggregated over the test splits described above.

\textbf{Auxiliary LLM.}
VQG and semantic consistency checking use an auxiliary text LLM in a text-only mode. Our primary choice is DeepSeek-V3.2-Exp~\cite{deepseekai2024deepseekv32} (zero sampling temperature for determinism). We also report results with GPT-4o~\cite{hurst2024gpt} and Gemini-2.5-flash~\cite{comanici2025gemini} for demonstrating the robustness of verification question generator choice in a controlled ablation (see Table~\ref{tab:llm}).

\subsection{Comparison Results}
Table~\ref{tab:main_result} reports AUC and AUG for all detection methods across datasets and model backbones. 
Generally, \textit{\textbf{V-Loop achieves the highest $\text{AUC}/\text{AUG}$ scores}} in the majority of settings, demonstrating particularly large gains on the \textit{\textbf{open-ended subsets}}.
The gains are smaller on the all splits because closed-ended samples often lack the dual semantic units required for logic-based verification.
V-Loop therefore falls back to the weaker rephrase strategy, behaving more like standard consistency methods.
Table \ref{tab:complement} shows that integrating V-Loop consistently brings substantial gains across all uncertainty-based baselines (SE, RadFlag, and VASE) and all MLLMs.
These improvements arise because uncertainty-based methods assess predictive uncertainty over the input pair, while V-Loop directly verifies the factual correctness of each specific answer.
By combining global uncertainty cues with explicit answer-level verification, the enhanced variants deliver stronger and more reliable hallucination detection.

\begin{table}
  \centering
  \caption{
  Performance (AUC (\%) and AUG (\%)) of V-Loop and its variants on the open-ended splits of VQA-RAD and VQA-Med-2019 datasets with Lingshu-7B.
  VQG and VAC are the Verification Question Generation and Visual Attention Consistency Enforcement, respectively.
  }
  \setlength\tabcolsep{6pt}
  \begin{tabular}{c|c|c|c|c|c}
    \toprule
    \multicolumn{2}{c|}{Components } & \multicolumn{2}{c|}{VQA-RAD} & \multicolumn{2}{c}{VQA-Med-2019} \\
    \hline
    \makecell[c]{VQG} & \makecell[c]{VAC} & AUC & AUG & AUC & AUG  \\
    \bottomrule
    \usym{2717} & \usym{2717} & 50.00 & 49.10 & 50.00 & 49.76 \\
    \usym{2717} & \usym{2713} 
    & 56.89\color{ForestGreen}$\uparrow$ & 50.06\color{ForestGreen}$\uparrow$ 
    & 55.04\color{ForestGreen}$\uparrow$ & 40.99\color{Red}$\downarrow$ \\
    \usym{2713} & \usym{2717} 
    & 66.79\color{ForestGreen}$\uparrow$ & 56.24\color{ForestGreen}$\uparrow$ & 69.98\color{ForestGreen}$\uparrow$ & 64.16\color{ForestGreen}$\uparrow$ \\
    \usym{2713} & \usym{2713} 
    & \textbf{70.00}\color{ForestGreen}$\uparrow$ & \textbf{56.47}\color{ForestGreen}$\uparrow$ 
    & \textbf{71.93}\color{ForestGreen}$\uparrow$ & \textbf{66.33}\color{ForestGreen}$\uparrow$ \\
    \bottomrule
  \end{tabular}
  \label{tab:ablation_module}
\end{table}

\begin{figure}[t]
  \centering
   \includegraphics[width=\linewidth]{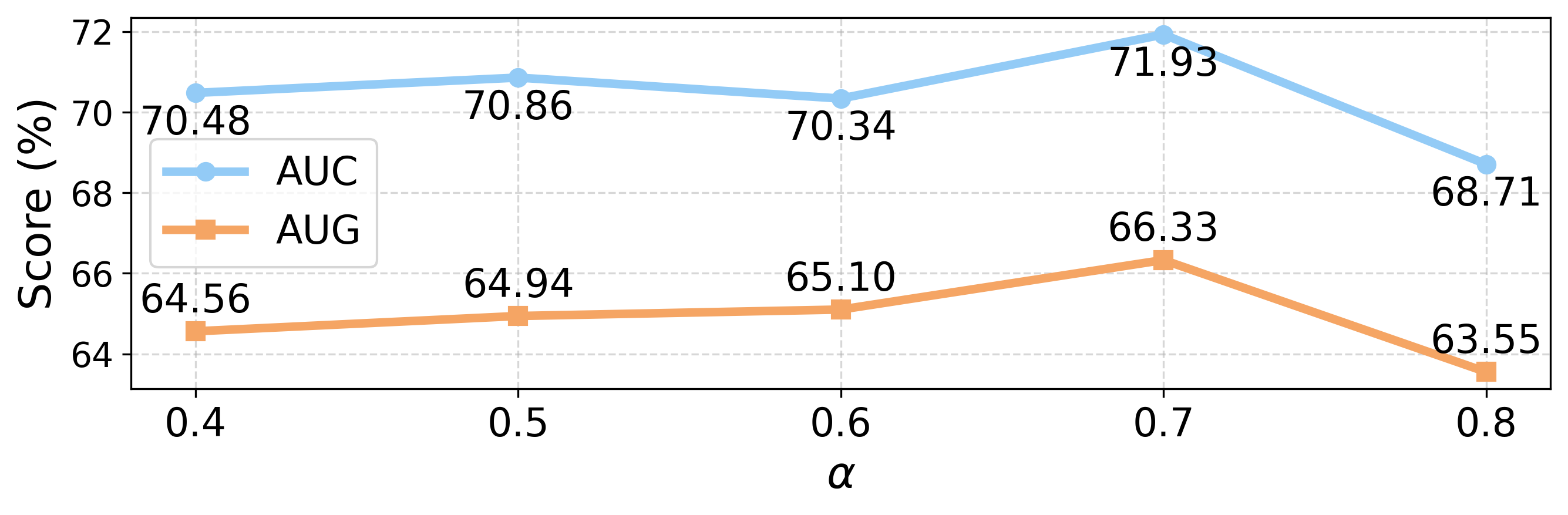}
   \caption{
   Effect of the weighting factor $\alpha$ on the performance (AUC (\%) and AUG (\%)) of V-Loop, evaluated on the VQA-RAD open-ended subset with MedGemma-4B-it. 
   }
   \label{fig:alpha}
\end{figure}

\begin{table}
  \centering
  \caption{
  Performance (AUC(\%) and AUG(\%)) of V-Loop variants using different auxiliary LLMs for verification question generation on the VQA-RAD open-ended subset with Lingshu-7B.
  }
  \setlength\tabcolsep{12pt}
  \begin{tabular}{l|c|c}
    \toprule 
    Auxiliary LLM & AUC & AUG \\
    \bottomrule
    DeepSeek-V3.2-Exp~\cite{deepseekai2024deepseekv32} & 70.00 & 56.47 \\
    GPT-4o~\cite{menick2024gpt} & 68.81 & 58.35 \\
    Gemini-2.5-Flash~\cite{comanici2025gemini} & 64.18 & 50.41 \\
    \bottomrule
  \end{tabular}
  \label{tab:llm}
\end{table}

\begin{figure*}[t]
  \centering
   \includegraphics[width=0.94\linewidth]{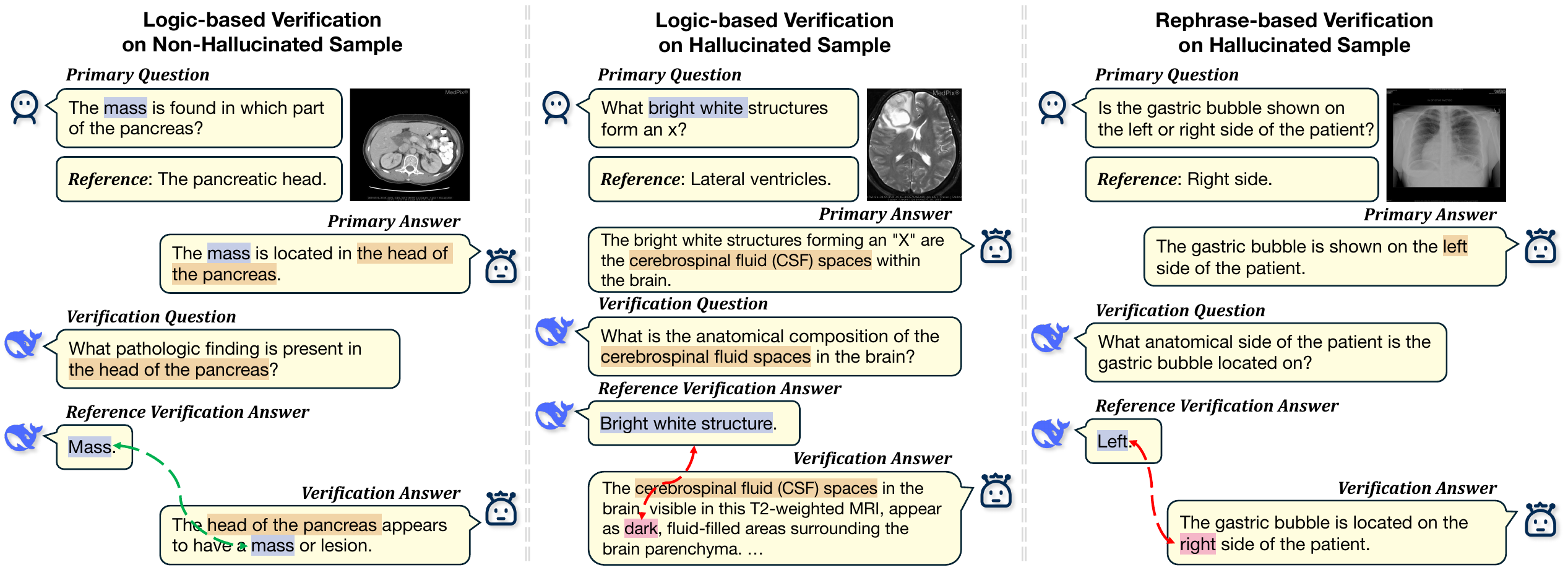}
   \caption{
   Examples of V-Loop.
   In each example, semantic units extracted from the question (blue) and answer (pink) are used to form the verification question. A correct verification answer closes the loop (left), while mismatches indicate hallucination (middle/right).
   }
   \label{fig:sample}
\end{figure*}

\subsection{Ablation Study}

\textbf{Module ablation.}  
We performed an ablation study to evaluate the contribution of two core components in V-Loop: VQG and VAC. We consider two ablated variants:  
(i) \textit{w/o VQG}: the VQG module is removed and the original dataset questions are reused as $q_{vri}$ with the generation temperature being set to 1.0 to encourage response diversity; and
(ii) \textit{w/o VAC}: the visual attention consistency constraint is disabled during answering the verification question.  
The ablation results presented in Table~\ref{tab:ablation_module} consistently show that removing either the VQG module or the VAC mechanism consistently degrades AUC and AUG, thereby confirming their essential and complementary roles.
The VQG module materially improves the diagnostic power by providing semantically targeted verification questions that enable effective reasoning, while the VAC mechanism stabilizes the visual grounding during this process by enforcing the reuse of original visual evidence, which is crucial for reducing false positives. 
Overall, the complete V-Loop framework achieves the best results, highlighting their necessity for reliable, visually grounded hallucination detection.

\begin{figure*}[t]
  \centering
  \begin{subfigure}[t]{0.45\textwidth}
    \centering
    \includegraphics[width=\linewidth]{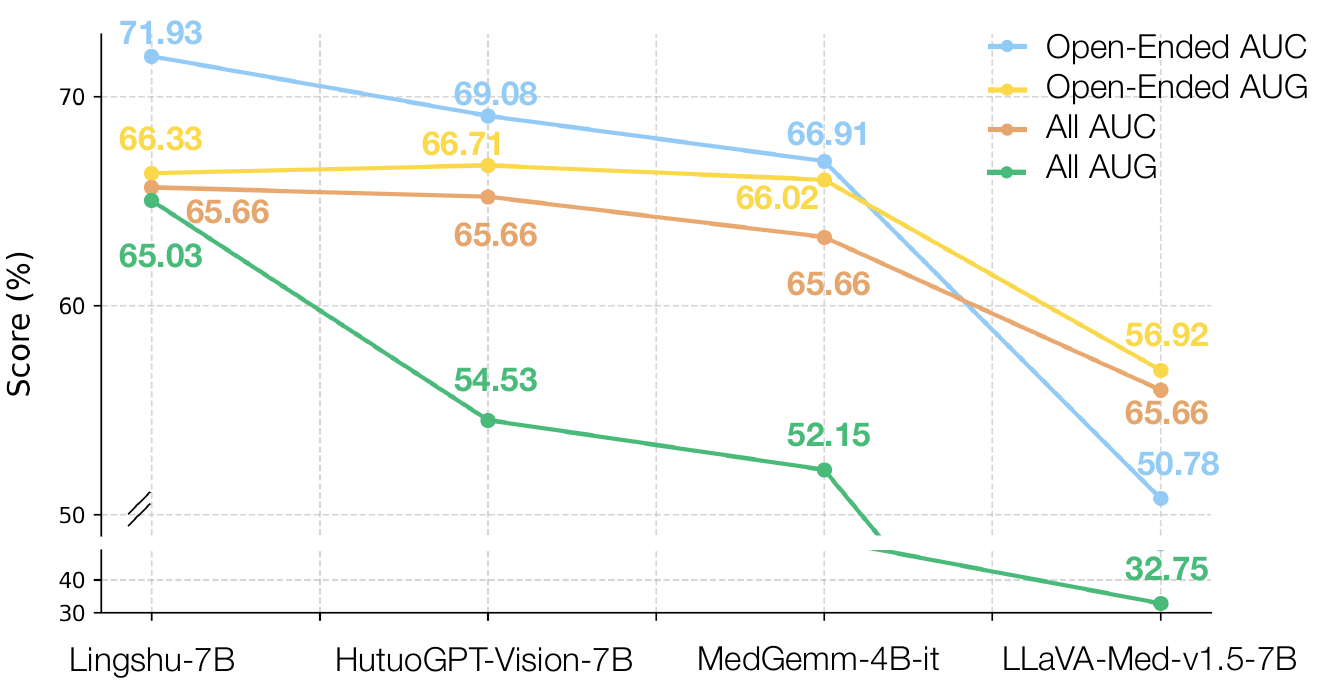}
    \caption{Performance (AUC (\%) and AUG (\%)) of V-Loop on VQA-Med-2019 across medical MLLMs.}
    \label{fig:dependency-a}
  \end{subfigure}
  \hfill
  \begin{subfigure}[t]{0.45\textwidth}
    \centering
    \includegraphics[width=\linewidth]{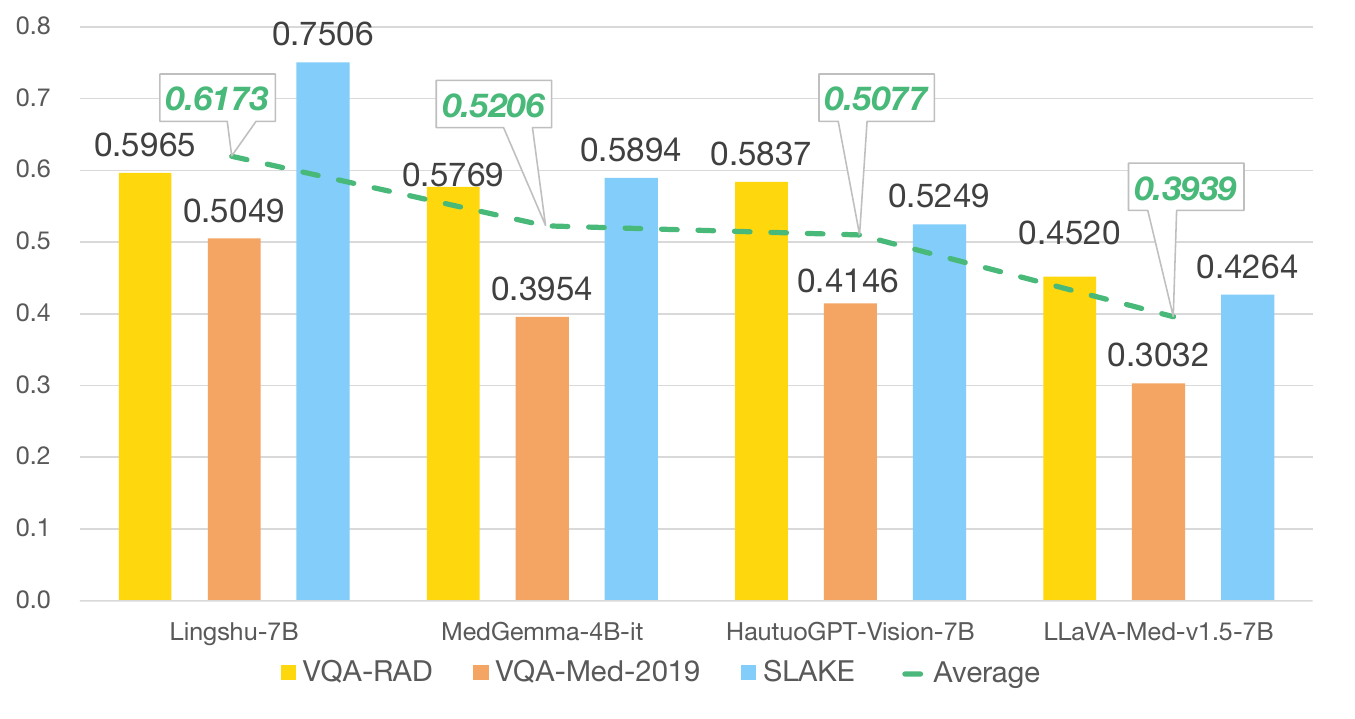}
    \caption{Performance of medical MLLMs measured by the average GREEN score across all datasets.}
    \label{fig:dependency-b}
  \end{subfigure}
  \caption{
  Correlation between MLLM capability and V-Loop detection performance. Stronger models (higher GREEN scores) deliver higher AUC/AUG with V-Loop, highlighting its dependence on the MLLM’s visual–semantic reasoning quality.
  }
  \label{fig:dependency}
\end{figure*}

\textbf{Impact of LLM Quality on VQG.}
Table~\ref{tab:llm} examines the impact of using different auxiliary LLMs to generate verification questions. Across all settings, V-Loop consistently improves detection performance over the baseline, indicating that the framework is robust to the choice of auxiliary model. Stronger LLMs (e.g., DeepSeek-V3.2-Exp) yield larger gains, as they produce more accurate and clinically coherent verification questions, while smaller models (e.g., Gemini-2.5-Flash) still lead to substantial improvements. This shows that V-Loop is effective regardless of the auxiliary LLM used, and benefits further when higher-quality verification questions are available.

\textbf{Hyperparameter Selection.}  
The weight factor $\alpha$ in Eq.~\ref{eq:attn_weight_reweight} controls the strength of the visual attention consistency constraint by scaling the aggregated attention $\bar{A}_{t \rightarrow v}$.  
Fig.~\ref {fig:alpha} presents the performance of V-Loop with different values of $\alpha$ on the VQA-RAD open-ended split. 
The proposed V-Loop framework maintains stable performance across a wide range of $\alpha$ values, demonstrating its robustness to parameter variation.  
The best results are achieved at $\alpha = 0.7$, where the model attains the optimal balance between maintaining visual grounding and allowing adaptive attention shifts during verification.

\subsection{Impact of MLLM Competence}
V-Loop relies on the underlying MLLM’s visual grounding and reasoning ability, making it sensitive to model quality.
In Fig.~\ref{fig:dependency-a}, V-Loop achieves notably lower AUC and AUG on weaker models such as LLaVA-Med-v1.5-7B~\cite{li2023llava}.
To quantify this dependency, we compute each model’s average GREEN score across all datasets (Fig.~\ref{fig:dependency-b}), where higher scores indicate stronger visual–semantic alignment.
Models with low GREEN scores (e.g., LLaVA-Med-v1.5-7B) exhibit substantial hallucinations in both primary and verification stages, reducing V-Loop’s effectiveness, while stronger models (e.g., Lingshu-7B, MedGemma-4B-it) enable more accurate logical-loop verification.
\section{Conclusion}

In this paper, we presented V-Loop, a Visual Logical Loop Verification framework for hallucination detection in medical VQA.
By constructing a verification question conditioned on the primary question–answer pair and constraining the model to answer it within the same visual region attended during the primary response, V-Loop enables the MLLM to detect hallucinated answers in a training-free, plug-and-play manner.
Extensive experiments on multiple medical VQA benchmarks and MLLMs demonstrate that V-Loop outperforms other introspective detection methods and improves uncertainty-based methods when combined.
{
    \small
    \clearpage
    \bibliographystyle{ieeenat_fullname}
    \bibliography{main}
}

\end{document}